\documentclass[10pt,twocolumn,letterpaper]{article}

\usepackage{wacv}
\usepackage{times}
\usepackage{epsfig}
\usepackage{graphicx}
\usepackage{amsmath}
\usepackage{amssymb}
\usepackage{multirow}
\usepackage{subcaption}

\def\wacvPaperID{735} 

\wacvfinalcopy 

\ifwacvfinal
\usepackage[breaklinks=true,bookmarks=false]{hyperref}
\else
\usepackage[pagebackref=true,breaklinks=true,colorlinks,bookmarks=false]{hyperref}
\fi

\begin{document}

\title{Siamese Transformer Pyramid Networks for Real-Time UAV Tracking}
\author{Daitao Xing \\
~~~~~~~~New York University , USA~~~~~~~~~     \\
{\tt\small daitao.xing@nyu.edu}
\and
Nikolaos Evangeliou\\
New York University Abu Dhabi, UAE\\
{\tt\small nikolaos.evangeliou@nyu.edu}
\and
Athanasios Tsoukalas\\
New York University Abu Dhabi, UAE\\
{\tt\small athanasios.tsoukalas@nyu.edu}
\and
Anthony Tzes\\
New York University Abu Dhabi, UAE\\
{\tt\small anthony.tzes@nyu.edu}
}

\maketitle
\thispagestyle{empty}

\begin{abstract}
Recent object tracking methods depend upon deep networks or convoluted architectures.
Most of those trackers can hardly meet real-time processing requirements on mobile platforms with limited computing resources. In this work, we introduce the Siamese Transformer Pyramid Network (SiamTPN), which inherits the advantages from both CNN and Transformer architectures. Specifically, we exploit the inherent feature pyramid of a lightweight network (ShuffleNetV2) and reinforce it with a Transformer to construct a robust target-specific appearance model. A centralized architecture with lateral cross attention is developed for building augmented high-level feature maps. To avoid the computation and memory intensity while fusing pyramid representations with the Transformer, we further introduce the pooling attention module, which significantly reduces memory and time complexity while improving the robustness. Comprehensive experiments on both aerial and prevalent tracking benchmarks achieve competitive results while operating at high speed, demonstrating the effectiveness of SiamTPN. Moreover, our fastest variant tracker operates over 30 Hz on a single CPU-core and obtaining an AUC score of $58.1\%$ on the LaSOT dataset. Source codes are available at \href{https://github.com/RISC-NYUAD/SiamTPNTracker}{https://github.com/RISC-NYUAD/SiamTPNTracker}
\end{abstract}
%
\section{Introduction}
\begin{figure}
   \begin{center}
   \includegraphics[width=0.47\textwidth]{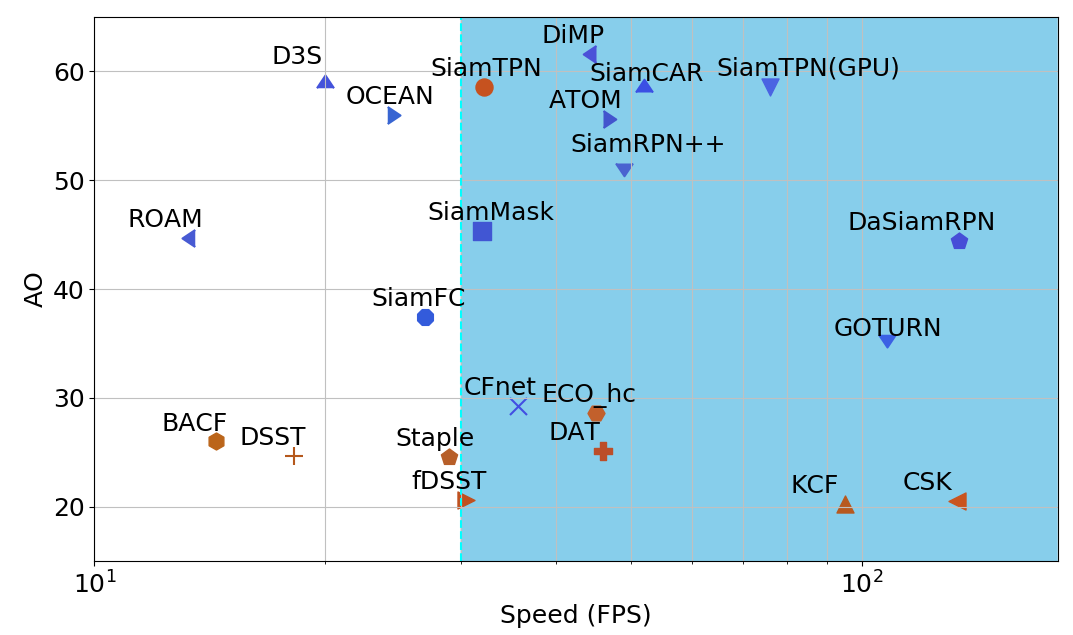}
   \end{center}
      \caption{A comparison of the quality and the speed on CPU (dark red) or GPU (blue) of tracking methods on Got10K test set. The Average Overlap (AO) with respect to the Frames-Per-Seconds (FPS) is presented. The blue area corresponds to the trackers running in real-time speed (above 30 FPS).}
   \label{got10k_ao_speed}
   \end{figure}
Unmanned Aerial Vehicle (UAV) tracking has drawn increasing attention in recent years given its enormous potential in diverse fields such as path planning~\cite{pathplanning}, visual surveillance~\cite{surveillance}, and border security~\cite{Tsoukalas}. While extensive advancements have been made towards powerful visual object tracking methods, 
the problem of real-time tracking has been overlooked. Moreover, the inherently limited power resources on lower performance compact devices further constraint the development of UAV tracking. \\
\indent{}Due to the optimization of both software and hardware on mobile devices and the progress of the lightweight but powerful backbone networks~\cite{alexnet, shufflenetv2, mobilenetv2}, the real-time applications based on visual classification, object detection, instance segmentation have been implemented on the CPU end. However, designing an efficient and effective object tracker for UAVs with limited computing resources, such as a single CPU-core, remains challenging. The lightweight backbones are insufficient for extracting robust discriminative features, which is vital for the tracking performance, especially under uncertainty scenarios. Thus, previous trackers try to address this problem by employing deeper networks~\cite{siamrpn++}, designing complex structures~\cite{ocean}, or online updaters~\cite{dimp}, which sacrifice the inference speed.\\
\begin{figure*}
\begin{center}
\includegraphics[width=0.9\textwidth]{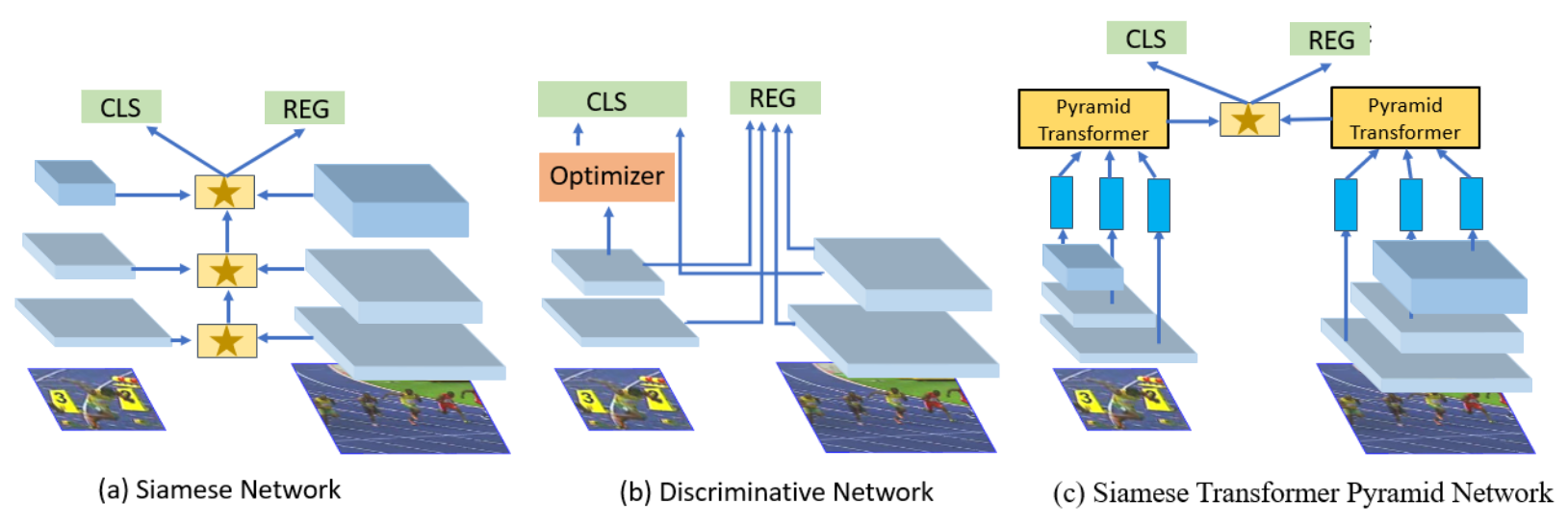}
\end{center}
   \caption{Object tracking architecture comparison. (a) Siamese based tracking network which operates cross correlation on pyramid layers separately. (b) Discriminative network that uses pyramid features for different tasks. (c) The proposed SiamTPN where features are first fused by the pyramid transformer module before used for both classification and regression.}
\label{framework}
\end{figure*}
\indent{}In this work, we alleviate the aforementioned problems, accommodate the lightweight backbone and build a real-time CPU-based tracker. Firstly, to complement the representative ability of lightweight backbone network, we integrate the Feature Pyramid Network (FPN)~\cite{fpn} into the tracking pipeline. Although existing trackers~\cite{atom, siamcar, siamrpn} also employ multi-scale features, most of them resort to a simple combination or use features for different tasks. We claim that this is fundamentally limited since a discriminative representation requires combining the contexts from multiple scales. Even though, FPN encodes the pyramid information from low/high level semantics, it only exploits contexts from local neighborhoods rather than explicitly modeling the global interactions. The perception of the FPN is constrained by the receptive field, which is limited on the shallower networks. Inspired by the development of Transformer~\cite{detr} and its  ability to model global dependencies, recent works~\cite{cgacd, siamatt} introduce attention-based modules and achieve profound results. However, the complexity of these models may cause computation/memory overhead which is not suited for pyramid architecture. Instead, we design a lightweight Transformer attention layer and embed it into pyramid network. The proposed Siamese Transformer Pyramid Network (named SiamTPN) augments the target features with lateral cross attention between pyramid features, producing robust target-specific appearance representation. Figure \ref{framework} illustrates the main difference between our tracker and existing ones. Moreover, our tracker based on a lightweight backbone network achieves state-of-the-art results while running at real-time speed on both GPU and CPU end, as shown in Figure \ref{got10k_ao_speed}. Our main contributions are summarized as follows:
\begin{enumerate}
    \item We introduce a novel Transformer-based tracking framework for systems with limited computational resouces. These systems are typical encountered in UAVs with only CPU-support. To the best of our knowledge, this is the first deep learning based visual tracker running at real-time speed on UAVs using CPUs.
    \item We propose a lightweight Transformer layer and integrate it into pyramid networks to build an efficient and effective framework.
    \item  Superior performance on multiple benchmarks as well as extensive ablation studies demonstrate the effectiveness of the proposed method. Particularly, our approach achieves state-of-the-art results and an AUC score of 58.1 on LaSOT~\cite{lasot} with only a lightweight backbone while running at over 30 FPS on the CPU end. The field tests further validate the efficiency of SiamTPN in real world applications.
\end{enumerate}
\section{Related Work}
\subsection{Lightweight Network}
With the requirement of running neural networks on mobile platforms, a series of lightweight models are proposed~\cite{alexnet, shufflenetv2, mobilenetv2}. AlexNet~\cite{alexnet} utilizes fully convolutional operations and achieves profound results on ImageNet~\cite{imagenet} classification tasks. MobileNet~\cite{mobilenetv2} family proposes inverted residual block, depthwise separate convolution to save computation cost. The ShuffleNet~\cite{shufflenetv2} family is another series of lightweight deep neural networks which introduce channel shuffle operation and optimize the network design for the target hardware.\\
\textbf{Feature Pyramid Network}
The feature pyramid (i.e. bottom-up feature pyramid) is the most common architecture in modern neural network design. The hierarchical structure of CNN encodes the contexts in the gradually increased receptive field. The Feature Pyramid Network (FPN)~\cite{fpn} and Path Aggregation Network (PANet)~\cite{panet} are commonly used for the cross-scale feature interaction and multi-scale feature fusion. FPN includes a bottom-up as well as a top-down path to propagate semantic information into multi-level features.
\subsection{Object Tracking}
\textbf{Discriminative correlation filter (DCF).} DCFs have shown promising results for object tracking since MOSSE~\cite{mosse} and KCF~\cite{kcf}. After that, multi-channel features, color names and multi scale features are used~\cite{dsst, dat} to improve the tracking robustness. Further improvements are achieved with non-linear kernels~\cite{SRDCF, STRCF}, long-term memory~\cite{ECO} and deep features~\cite{ccot, goturn}. \cite{arcf, autotrack} further improves the robustness and optimized DCF for UAV tracking.\\
\textbf{Deep learning based object tracking} 
The popular Siamese network family based trackers address object tracking via similarity learning. SiamRPN~\cite{siamrpn} introduces the region proposal network to jointly perform classification and regression. DaSiamRPN~\cite{dasiamrpn} improves the discrimination power of the model with a distractor-aware module and SiamRPN++~\cite{siamrpn++} further improves the performance with more powerful deep architectures. Recent works like SiamBAN~\cite{siamban1}, SiamFC++~\cite{siamfc++} and Ocean~\cite{ocean}  replace the RPN with an anchor-free mechanism and achieve faster tracking speed. DiMP~\cite{dimp} and ATOM~\cite{atom} learn a discriminative classifier online to distinguish the target from the background. These methods require intense calculation which is not suitable for CPU-based tracking. \\
\textbf{Transformer.} Transformer was first proposed for machine translation in~\cite{transformer} and shows great potential in many sequential tasks. DETR~\cite{detr} first migrated Transformer into object detection tasks and achieves remarkable results. Recent works~\cite{cgacd, siamatt} introduce an attention mechanism for improving the tracking performance. Motivated by DETR, \cite{hift} make use of transformer to directly fuse correlation maps from different levels and obtains remarkable accuracy and speed for object tracking on UAVs. Instead of migrating the complex transformer encoder and decoder paradigm, in this work, we exploit the transformer encoder and design an attention based feature pyramid fusion network to learn the target-specific model more efficiently. 
\section{Proposed Method}
%
As shown in Figure \ref{framework}, the proposed SiamTPN, consists of three modules: one Siamese backbone network for feature extraction, one Transformer based feature pyramid network and one prediction head for per-pixel classification and regression. 
\subsection{Feature Extraction Network}
Similar to a Siamese tracking framework, the proposed SiamTPN consists of two branches: the template branch, which takes cropped image $z$ of size $W_z \times H_z$ from the initial frame as reference, and the search branch, which takes the cropped image $x$ of size $W_x \times H_x$ from the current frame for tracking. The two inputs are processed by the same backbone network, obtaining pyramid feature maps $P_{i} \in \mathbb{R}^{C_{i} \times \frac{W}{R} \times \frac{H}{R}}$, where $i \in \{3,4, 5\}$ is the stage number of feature extraction and $R$ is spatial reduction ratios, $R_i \in \{8, 16, 32\}$.\\ 
\indent{}
Instead of performing cross-correlation directly on feature maps pairs, we first feed the feature pyramid into the TPN (details in Section 3.3), shared between template branch and search branch. Specifically, TPN takes pyramid features $P_3, P_4, P_5$ as input and output the blend representation with the same size of $P_4$ for correlation purpose. Then, a depth-wise correlation is performed between outputs from reference branch and search branch as:
\begin{equation}
M = \Gamma(P_3^x, P_4^x, P_5^x) \star \Gamma(P_3^z, P_4^z, P_5^z)~,
\end{equation}
where $\Gamma$ is the TPN module, and $M$ is a multi-channel correlation map and is adopted as the input to the classification and regression head. The overall architecture is shown in Figure \ref{framework}.\\
\subsection{Feature Fusion Network}
\begin{figure}
\begin{center}
\includegraphics[width=0.50\textwidth]{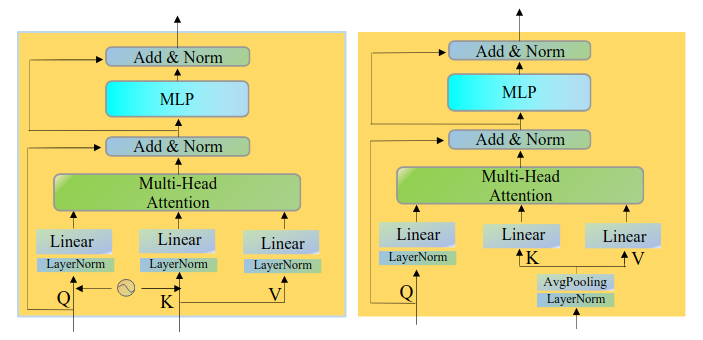}
\end{center}
   \caption{\textbf{Multi-head attention module versus Pooling Attention (PA) module.} Compared with the original attention block, the memory and time complexity in PA module is independent to the size of input features and controlled by the pooling operation}
\label{PAmodule}
\end{figure}

\noindent{}\textbf{Multi-head Attention.} Generally, a Transformer has several encoder layers, and each encoder layer is composed of Multi-head attention (MHA) module and a multilayer perceptron (MLP) module. The attention function is operated on queries $\mathbf{Q}$, keys $\mathbf{K}$ and values $\mathbf{V}$ in the scale dot-production way, which can be expressed as:
\begin{equation}
\begin{gathered}
\text { Attention }(\mathbf{Q} , \mathbf{K}, \mathbf{V})= \\ \operatorname{softmax}\left(\frac{\mathbf{(Q+Pos) (K+Pos)}^{\top}}{\sqrt{C}}\right) \mathbf{V}
\end{gathered}
\end{equation}
where the $C$ is the key dimensionality to normalize the attention , and $\mathbf{Pos}$ is the positional encoding that are added to the input of each attention layer. The positional embedding in Transformer architectures is a location-dependent trainable parameter vector that is added to the token embeddings prior to inputting them to the Transformer blocks. The model representation capability is enhanced when extending the attention mechanism into multiple head way, which can be formulated as follows:
\begin{equation}
\hspace*{-4mm}\operatorname{MultiHead}(\mathbf{Q}, \mathbf{K}, \mathbf{V})=\operatorname{Concat}\left(\mathbf{H}_{1}, \ldots, \mathbf{H}_{N}\right) W^{O},
\end{equation}
\begin{equation}
\mathbf{H}_{i}=\operatorname{Attention}\left(\mathbf{Q} W_{i}^{Q}, \mathbf{K} W_{i}^{K}, \mathbf{V} W_{i}^{V}\right),
\end{equation}
where $W_{i}^{Q} \in \mathbb{R}^{C \times d_{head}}, W_{i}^{K} \in \mathbb{R}^{C \times d_{\text {head }}}, W_{i}^{V} \in \mathbb{R}^{C \times d_{\text {head }}}$ and $W^{O} \in \mathbb{R}^{C \times C}$ are parameters of linear projections, $\operatorname{Concat}$ refers to concatenation operation, $N$ is the number of attention head, and $d_{head}$ is the dimension of each head equal to $\frac{C}{N}$. \\
\textbf{Pooling Attention.} MHA made the model assign importance to different aspects of information and learns a robust representation. However, the complexity increases with the power of input size. The computational cost for MHA is:
\begin{equation}
\mathcal{O}(MHA) = 2 \times n_q n_{kv} C+n_q C^{2} + n_{kv} C^{2},
\end{equation}
where $n_q = h_q w_q,  n_{kv} = h_{kv} w_{kv}$, $w, h$ is the resolution of input feature map. There exist three ways to reduce the computation cost: (1) reduce the query size, (2) reduce the dimension of $C$, or (3) reduce the key and value size. However, reducing the query size also reduces the number of points for the prediction head, which eventually affects tracking accuracy. The same situation happens with the reduction of feature dimensionality. Since the feature maps with variable resolution are used as keys and values for fusion in TPN, we propose a pooling attention (RA) layer to reduce the spatial scale of $\mathbf{K}$ and $\mathbf{V}$. Specifically, the $\mathbf{K}$ and $\mathbf{V}$ are fed into a pooling layer with both pooling and stride size of $R$. \\ 
\indent{}To further reduce the computation cost of attention module, we remove the position encoding in original MHA for the following reasons: (1) the permutation of the input tokens is constrained by the final cross correlation. (2) Accessing and storage of the position embedding for each feature maps costs extra resources which is not suited for mobile devices. Overall, the mechanism of PA block (PAB) can be summarized as:
\begin{equation}
\operatorname{PAB}(\mathbf{Q}, \mathbf{K}, \mathbf{V}, R) = \operatorname{Norm} (\mathbf{F} + \operatorname{MLP}\left(\mathbf{F}\right)),
\end{equation}
\begin{equation}
\mathbf{F} =\operatorname{Norm}(\mathbf{Q}+\operatorname{PA}(R)\left(\mathbf{Q}, \mathbf{K}, \mathbf{V}\right)),
\end{equation}
where $\operatorname{MLP}$ is a fully connected feed-forward network, and $\operatorname{Norm}$ is the LayerNorm to smooth the input feature. The structure comparison between MHA and PA module is shown in Figure \ref{PAmodule}.
\subsection{Transformer Pyramid Network }
\begin{figure}[htbp]
\begin{center}
\includegraphics[width=0.50\textwidth]{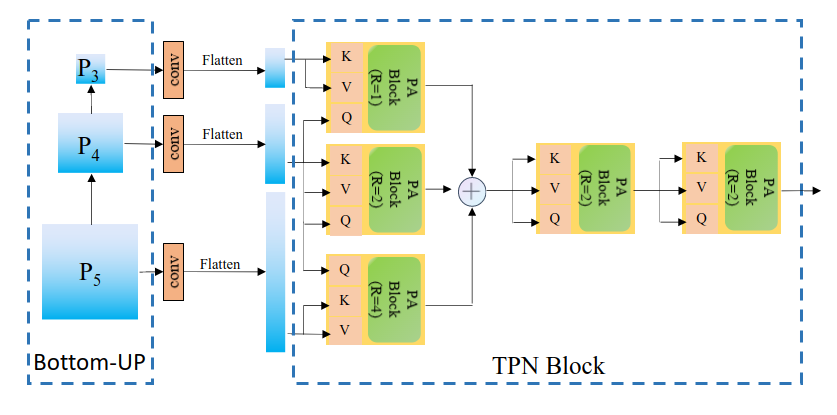}
\end{center}
   \caption{\textbf{Transformer Pyramid Network (TPN).}~Features from different levels $P_3-P_5$ are flattened and fed into TPN blocks. Each TPN block consists of 5 PA layers. Hierarchical information are extracted by 3 separate PA layers and further distilled by 2 additional PA layers. Variable stride pooling ratios $R$ are assigned for each layer for efficiency purposes.}
\label{TPNmodule}
\end{figure}

To leverage the pyramid feature hierarchy $P_i,~i \in\{3, 4, 5\}$, which has both low-level information and high-level semantics, a Transformer Pyramid Network (TPN) is proposed to build a blend feature with high-level semantics throughout. The TPN consists of stacked TPN blocks which takes pyramid features $\{P_3, P_4, P_5\}$ and output new fusion feature $\{P_3^{'}, P_4^{'}, P_5^{'}\}$, as shown in Figure \ref{TPNmodule}. The pyramid features are fed into a $1\times1$ convolution layer for dimension reduction, following a flatten operation before processing in the TPN. We fix the feature dimension (numbers of channels), denoted
as $C$ in all the feature maps.\\
\indent{}The construction of the pyramid features involves a bottom-up pathway and centralized pathway. The bottom-up pathway is the feed-forward convolution from the backbone architecture and produces feature hierarchy $\{P_3, P_4, P_5\}$. Then a centralized pathway merges the feature hierarchy into a unified feature. Specifically, we use $P_4$ as query for all feature hierarchy, yielding 3 combinations with different pooling scales which are processed by three parallel PAB locks. The outputs are directly added and fed into two self-attention PAB blocks to get the final semantic feature. The whole processing can be formulated as:
\begin{equation}
\begin{gathered}
P_4^{'} = \operatorname{PAB}(P_4,P_3,P_3,R=4) + \\
\operatorname{PAB}(P_4,P_4,P_4,R=2)  + \operatorname{PAB}(P_4,P_5,P_5,R=1)
\end{gathered}
\end{equation}
\begin{equation}
P_4^{'} = \{\operatorname{PAB}(P_4^{'},P_4^{'},P_4^{'},R=2)\}_{n=2}
\end{equation}
\begin{equation}
P_5^{'} = P_5 ; P_3^{'} = P_3.
\end{equation}
$P_3$ and $P_5$ are set as identity ones to avoid computation/memory overhead. Moreover, PA block design guarantee that the interdependencies among hierarchical features can be raised efficiently. The TPN Block repeats B times and produces the final representation for cross-correlation and the final prediction. Simplicity is central to our design and we have found that our model is robust to various design choices.
\subsection{Prediction Head}
%
The fusion features $P_4^x$ and $P_4^z$ are reshaped back to the original size before fed into the prediction head. Following~\cite{siamrpn++}, the Depth-wise Cross Correlation is performed between the search map and the template kernel to get a mult-channel correlation map. The correlation maps are fed into two separate branches. Each branch consists of 3 stacked convolution blocks to generate final outputs $A_{w \times h \times 2}^{c l s}$ and $A_{w \times h \times 2}^{reg}$. $A_{w \times h \times 2}^{c l s}$ represents the foreground and background scores for each point on feature maps and $A_{w \times h \times 2}^{reg}$ predicts the distances from each feature point to the four sides of the bounding box. Overall, the objective function is 
\begin{equation}
\mathcal{L}=\lambda_{cls} \mathcal{L}_{c l s}+\lambda_{iou } \mathcal{L}_{iou}+\lambda_{reg} \mathcal{L}_{reg},
\end{equation}
where $\mathcal{L}_{c l s}$ is the cross-entropy loss for classification,  $\mathcal{L}_{iou}$ is GIOU~\cite{giou} loss between prediction boxes and ground truth box and $\mathcal{L}_{reg}$ is the $L1$ loss for regression. Constants $\lambda_{cls}$, $\lambda_{reg}$ and $\lambda_{iou}$ weight the losses.
\section{Experimental Studies}
This section first presents the implementation details and the comparisons between variants of the SiamTPN tracker, with the cross-correlation visualization results. Then, ablation studies are presented to analyze the effects of the key components. We further compare our method with the state-of-the-art methods both on aerial and prevalent benchmarks. Finally, we deployed our tracker on a UAV platform to test its effectiveness in real-world applications.
\subsection{Implementation Details}
\textbf{Model}
We apply our SiamTPN to three representative lightweight backbones, namely AlexNet~\cite{alexnet}, MobileNetV2~\cite{mobilenetv2}, ShuffleNetV2~\cite{shufflenetv2}. Using those networks as backbones enables us to adequately compare the effectiveness of proposed method. All backbones are pretrained on Imagenet. The details of backbone configuration for the different backbones are shown in Table \ref{backbone}. For ShuffleNet and MibileNet, we extract that the stages of spatial ratio equal to ${\frac{1}{8}, \frac{1}{16}, \frac{1}{32}}$ respectively. For AlexNet, the last three layers are used for building feature pyramid. \\
\begin{table}[htbp]
\centering
\resizebox{0.44\textwidth}{!}{%
\begin{tabular}{|c|cc|cc|cc|}
\hline
\multirow{2}{*}{Backbone} & \multicolumn{2}{c|}{AlexNet~\cite{alexnet}}   & \multicolumn{2}{c|}{MobileNet~\cite{mobilenetv2}} & \multicolumn{2}{c|}{ShuffleNet~\cite{shufflenetv2}} \\ \cline{2-7} 
                          & \multicolumn{1}{c|}{stage} & C & \multicolumn{1}{c|}{stage} & C & \multicolumn{1}{c|}{stage}  & C \\ \hline
$P_3$          & 3           & 384         & 2           & 32          & 2          & 116          \\
$P_4$          & 4           & 384         & 4           & 96          & 3          & 232          \\
$P_5$          & 5           & 256         & 6           & 320         & 4          & 464          \\ \hline
\#Param (M) & \multicolumn{2}{c|}{3.1}  & \multicolumn{2}{c|}{1.81}  & \multicolumn{2}{c|}{0.8} \\
GFLOPs      & \multicolumn{2}{c|}{4.33} & \multicolumn{2}{c|}{0.39} & \multicolumn{2}{c|}{0.16} \\ \hline
\end{tabular}%
}
\caption{Backbone configurations. $\#$Param refer to the number of parameters. Multi-Adds GFLOPS is calculated under the input size of $256 \times 256$ for feature extraction. C is the dimension of the stage.}
\label{backbone}
\end{table}

\textbf{Training}
Like the Siamese approaches, the network is trained offline with image pairs. The training data consists of the training splits from LaSOT~\cite{lasot}, GOT10K~\cite{got10k}, COCO~\cite{coco} and TrackingNet~\cite{trackingnet} dataset. The image pairs are sampled from the videos with a maximum gap of 100 frames. The sizes of search images and templates are 256 × 256 pixels and 80 × 80 pixels respectively, corresponding to $4^2$ and $1.5^2$ times of the target box area, resulting in pyramid features $\{h_3^x=h_3^x=32,h_4^x=h_4^x=16,h_5^x=h_5^x=8\}$ and $\{h_3^z=h_3^z=10,h_4^z=h_4^z=5,h_5^z=h_5^z=3\}$. Even though the lower input resolution brings additional speed increment, it is not the focus of this paper, so we set the aforementioned sizes for all the following experiments. The test images are augmented with some perturbation in the position and scale. \\
\indent{}For all backbones, the first layer and all BatchNorm layers are frozen during training. All experiments are trained for 100 epochs with 64 image pairs per batch. We use the ADAMW~\cite{adamw} optimizer with initial learning rate of $10^{-5}$ for the backbone and $10^{-4}$ for the rest of the parts. The learning rate drops by a factor 0.1 decay on 90 epochs and the loss terms are weights with $\lambda_{cls}=5$,$\lambda_{iou}=5$,$\lambda_{reg}=2$ respectively. During tracking, the scale penalty and Hanning windows~\cite{csk} is performed before selecting best prediction point from classification map $A_{w \times h \times 2}^{c l s}$. The final bounding box is given by adding the offsets predicted in $A_{w \times h \times 2}^{r e g}$ to the coordinates of the best prediction point.
\subsection{Ablation Study}
In this section, we verify the effectiveness of the proposed tracker from the following aspects: backbones choice, comparison with original Transformer and Convolution, the impact of TPN hyperparameters and the attention visualization. We follow the one-pass evaluation (Success and Precision) to compare different tracking configurations on the LaSOT~\cite{lasot} test set and report the Success (AUC) scores. LaSOT~\cite{lasot} is a large-scale long-term tracking benchmark which contains 280 videos for testing.  
\begin{figure}[htbp]
\begin{center}
\includegraphics[width=0.50\textwidth]{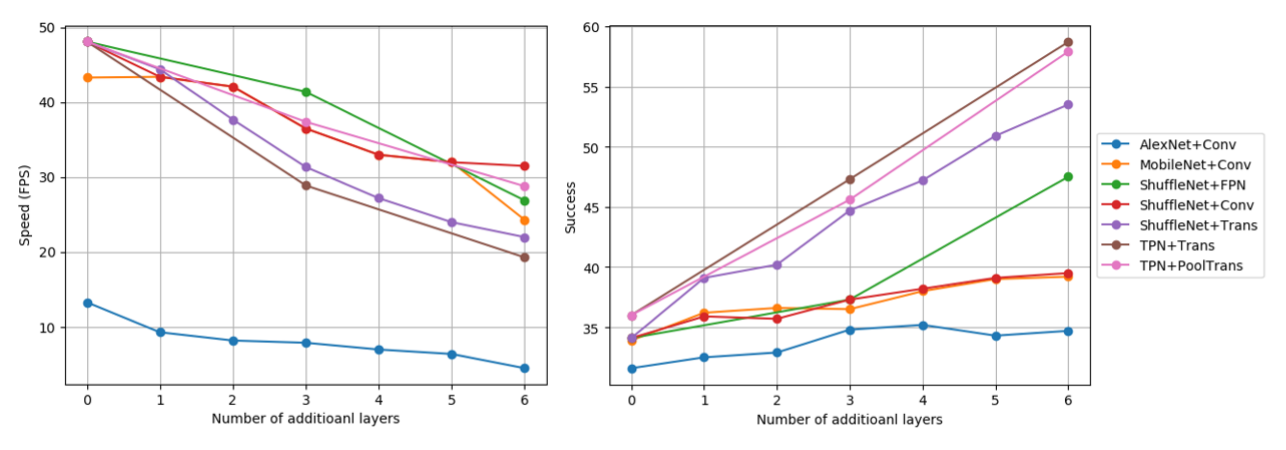}
\end{center}
   \caption{Speed and AUC score for different configurations.}
\label{ablation1fig}
\end{figure}

\begin{table*}[]
\centering
\resizebox{0.95\textwidth}{!}{%
\begin{tabular}{c|c|cccccccccc}
\hline
Backbone & Neck Type & $C$ & $N$ & B & Depth & AUC & $\Delta_{AUC}$ & \#Params (M) & GFLOPs & FPS (CPU) & $\Delta_{FPS}$ \\ \hline
\multirow{2}{*}{AlexNet~\cite{alexnet}}      & Identity & 192 &   &   & 0 & 31.6 & -     & 3.94  & 5.73 & 13.3 & \_    \\
                              & Conv     & 192 &   & 2 & 6 & 34.7 & +3.1  & 9.62  & 8.31 & 4.5  & -8.8  \\ \hline
\multirow{2}{*}{MobileNetV2~\cite{mobilenetv2}}  & Identity & 192 &   &   & 0 & 33.9 & +2.3  & 2.58  & 0.95 & 43.3 & +30.0 \\
                              & Conv     & 192 &   & 2 & 6 & 39.2 & +7.6  & 5.04  & 1.75 & 24.3 & +11.0 \\ \hline
\multirow{9}{*}{ShuffleNetV2~\cite{shufflenetv2}} & Identity & 192 &   &   & 0 & 34.1 & +2.5  & 1.57  & 0.6  & 48.1 & +34.8 \\
                              & Conv     & 192 &   & 2 & 6 & 39.5 & +7.9  & 3.56  & 1.4  & 31.2 & +17.9 \\ \cline{2-12} 
                              & FPN      & 192 &   & 2 & 6 & 47.5 & +15.9 & 3.85  & 1.62 & 26.9 & +13.3 \\
                              & Trans    & 192 & 6 & 2 & 6 & 53.5 & +21.9 & 4.24  & 1.79 & 22   & +8.7  \\
                              & TPNwoPA  & 192 & 6 & 2 & 6 & 58.7 & +27.1 & 4.84  & 2.05 & 17.7 & +4.4  \\ \cline{2-12} 
                              & TPN      & 192 & 6 & 1 & 3 & 52.8 & +21.2 & 3.92  & 1.08 & 33.2 & +19.9 \\
                              & TPN      & 128 & 4 & 2 & 6 & 46.2 & +14.6 & 2.7   & 0.88 & 37.1 & +23.8 \\
                              & TPN      & 192 & 6 & 2 & 6 & 58.1 & +26.5 & 4.24  & 1.31 & 32.1 & +18.8 \\
                              & TPN      & 256 & 8 & 2 & 6 & 58.4 & +26.8 & 10.77 & 3.73 & 15.2 & +1.9  \\ \hline
\end{tabular}%
}
\caption{\textbf{Comparison with different backbones and fusion configuration.} ``Identity" means no feature encoding between pyramid features and cross-correlation. ``Conv" and ``Trans" refer to the use of Convolution layer or original Transformer to encode features. In those cases, only $P_4$ is used since the is no pyramid information to merge. For ``TPN" and ``TPNwoPA" share the same setting except the PA blocks are replaced with the original Transformer. $C$ is feature channel, $N$ is the number of attention head and B is the repeat number of TPN blocks. The AUC score are tested on LaSOT~\cite{lasot} test set. Since a TPN block consists of 3 layers processing $P_4$, for fair comparison, each block in ``Conv" and ``Trans" represents 3 stacked corresponding layers. }
\label{ablation1}
\end{table*}

\begin{figure}[htbp]
\begin{center}
\includegraphics[width=0.47\textwidth]{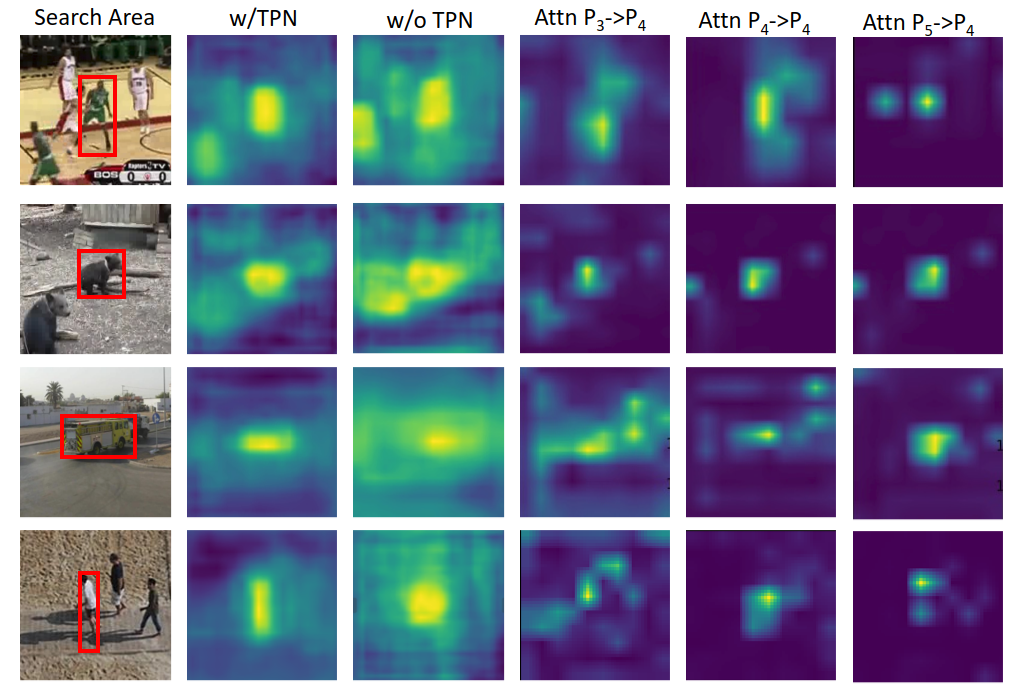}
\end{center}
   \caption{The visualization of response map with TPN (second column), without TPN (third column) and attention map between $P_i$ and $P_4$.}
\label{attn_visual}
\end{figure}

\noindent{}\textbf{Backbones.} The backbone network has the dominant impact on inference speed and accuracy. Modern architectures make use of residual skip connection, group/depth-wise convolution to design a competent network to learn more representative features, with even higher inference speed. We first compared the performance using different backbones. Similar to SiamFC~\cite{siamfc}, we remove all the feature fusion modules and predict results directly from $P_4$. We set $C$=192 for all prediction layers. As shown in Table \ref{ablation1}, the tracker with a simple backbone with prediction head achieves appreciable AUC scores on LaSOT with an average high inference speed on CPU end. Specifically, ShuffleNetV2 achieves AUC score of 34.1 with 48.1 FPS. A straight forward question is: Will more attached convolution layers help with tracking performance? We then stack additional convolution layers following $P_4$ and Figure~\ref{ablation1fig} shows the AUC changes along with the number of additional layers. Stacking more convolution layers improves the accuracy inefficiently and is worthless when compared with the speed drop. For ShufflenetV2, the speed drops over $30\%$ at a $15\%$ improvement on AUC score. We see that AlexNet is not suited for edge computing and ShuffleNetV2 and MobileNetV2 give comparable results both on accuracy and speed test. For the following experiments, we choose ShuffleNetV2 as the backbone.\\
\noindent{}\textbf{Comparison with original Transformer.}
To show the effect of our proposed TPN module and PA block, we design a tracker using the original Transformer. Similar to the setting of stacked convolution, we attach additional Transformer layers behind $P_4$. As shown in Figure~\ref{ablation1fig}, without the fusion of pyramid features, the tracker with only one additional transformer layer achieves better results than the tracker with six additional convolution layers. Moreover, the tracker with six transformer layers achieves an AUC score of 53.5 on LaSOT. 
Next, we implement an FPN using the same settings as TPN, but replacing the transformer layers with convolution and interpolation layers. The tracker with two stacked FPN learns more comprehensive representations from the interactions inside the feature pyramid and gets an AUC score of 47.2, demonstrating its advantage over the single layer architecture. However, the lack of the global dependencies become the bottleneck of improving accuracy. We further integrate Transformer layer into TPN blocks without using Pooling Attention layer. With the high-level semantics aggregated from pyramid features, the tracker achieves an state-of-the-art performance on LaSOT with an AUC score of 58.7. However, we see that the speed of tracker drops below 20 FPS which is not applicable for real-time tracking requirement. Finally, we test the results of TPN model with PA layers instead of transformer layer. Even the input size of queries and keys shrink with scale R, the tracker still achieves state-of-the-art performance . Nevertheless, the speed boosts up to 32.1 FPS with only 0.6 AUC score loss on LaSOT dataset, demonstrating the superiority of our method on both robustness and efficiency.\\
\begin{table*}[]
\centering
\resizebox{\textwidth}{!}{%
\begin{tabular}{c|cccccccccccccccc}
\hline
\multirow{2}{*}{Trackers} & KCF & BACF & CSRDCF & ARCF & Auto & ECO & Siam & DaSiam & HiFT & Siam & Siam & Siam & DiMP & ATOM & Siam & Siam \\
      & ~\cite{kcf}      & \cite{bacf}     & \cite{csrdcf}  & \cite{arcf}      & Track~\cite{autotrack} & \cite{ECO}     & RPN++~\cite{siamrpn++} & \cite{dasiamrpn}     & \cite{hift}     & BAN ~\cite{siamban1}  & CAR~\cite{siamcar}  & Attn~\cite{siamatt} &    ~\cite{dimp}  & ~\cite{atom}      & TPN  & TPN     \\ \hline
Feat  & HF   & HF   & HF   & HF   & HF    & VGG  & R50   & Alex & R50  & R50  & R50  & R50  & R50  & R18  & Alex & Shuffle \\ \hline
Prec. & 52.3 & 66.2 & 67.6 & 67.1 & 68.9  & 75.2 & 76.9  & 60.8 & 78.7 & 83.3 & 76   & \textcolor{blue}{84.5} & \textcolor{green}{84.9} & 83.7 & 79   & \textcolor{red}{85.83}   \\
Succ. & 33.1 & 46.1 & 48.1 & 46.8 & 47.2  & 52.2 & 57.9  & 40   & 58.9 & 63.1 & 61.4 & \textcolor{blue}{65}   & \textcolor{green}{65.4} & \textcolor{blue}{65}   & 59.3 & \textcolor{red}{66.04}   \\
FPS   & 95   & 14.4 & 58   & 15.3 & 65.4  & 45   & 35$^{*}$    & 134$^{*}$  & 130$^{*}$  & 40$^{*}$   & 52$^{*}$   & 45$^{*}$   & 45$^{*}$   & 46$^{*}$   & 105$^{*}$  & 32.1    \\ \hline
\end{tabular}%
}
\caption{Comparison results on UAV123 dataset~\cite{uav123} in terms of precision (Prec.), success (Succ.) and speed (FPS). HF refers to handcraft features, R50 (18) is Resnet-50 (18) \cite{resnet}, Alex, Shuffle, VGG represents AlexNet~\cite{alexnet}, ShuffleNet~\cite{shufflenetv2}, VGGNet~\cite{vgg} respectively. GPU speeds are mark with $^{*}$, Our SiamTPN based on AlexNet and ShuffleNet exhibit promising results. The top three trackers are shown in \textcolor{red}{red}, \textcolor{green}{green} and \textcolor{blue}{blue} fonts.}
\label{uav_result}
\end{table*}
\textbf{Impact of TPN hyperparameters.} We discuss some architecture hyper parameters of the TPN model. Firstly, we examine the impact of the number of TPN blocks. With only one TPN block, the tracker produces a slight speed increment but suffers from AUC score drop from 58.1 to 52.8. Since the original transformer use 6 layers depth for both the encoder and decoder, we argue that 2 TPN blocks (depth=6) are enough for achieving robust tracking results. The number of heads in PA layer also plays an important role in tracking stability. For simplicity, we fix the head dimension=32, so we can test the input dimension $C=\{128, 192, 256\}$ and head number $N=\{4,6,8\}$ simultaneously. The tracker with 8 heads yields the best AUC score albeit at a cost of reducing in half of the inference time (FPS from 32.1 to 15.2). On the other hand, only using 4 heads is inefficient to learn an effective representation and only gives an AUC score of 46.2 on LaSOT. In practice, $C$=192, $N$=6, B=2 gives best balance between speed and accuracy. \\
\begin{table}[]
\centering
\resizebox{0.47\textwidth}{!}{%
\begin{tabular}{c|ccccccclc}
\hline
    & Siam  & ATOM & Dimp & Siam & CGACD & SiamAttn & SiamBAN & Ocean & Ours \\
    & RPN++~\cite{siamrpn++} & \cite{atom} & \cite{dimp} & FC++~\cite{siamfc++} & ~\cite{cgacd} & ~\cite{siamatt}   & \cite{siamban1}       &  \cite{ocean}     &      \\ \hline
EAO & 41.4  & 40.1 & 44   & 42.6 & 44.9  & \textcolor{blue}{47.0}     & 45.2    & \textcolor{red}{48.9}  & \textcolor{green}{46.2} \\
AUC & 69.6  & 66.7 & 68.4 & 68.3 & \textcolor{red}{71.3}  & \textcolor{green}{71.2}     & 69.6    & 68.4  & \textcolor{blue}{71.0} \\ \hline
\end{tabular}%
}
\caption{Evaluation on VOT and OTB datasets}
\label{vot_otb}
\end{table}
\textbf{Attention Visualization.} The first three columns in Figure~\ref{attn_visual} show the response maps from the classification head with or without TPN module. Without TPN to learn discriminative features, the correlation results become dispersed and much easier to shift to distractors. The last three columns illustrate the attention maps between pyramid features. The attention between lower levels ($P_3$ to $P_4$, $P_4$ to $P_4$) distill more local information across the search area, while attention from high level ($P_5$ to $P_4$) is more centralized on the semantics of the object target. All attention maps are calculated from the central feature point inside the bounding box with the whole key inputs.

\subsection{Comparison with State-Of-The-Art Trackers}
\begin{figure}[htbp]
\begin{center}
\includegraphics[width=0.46\textwidth]{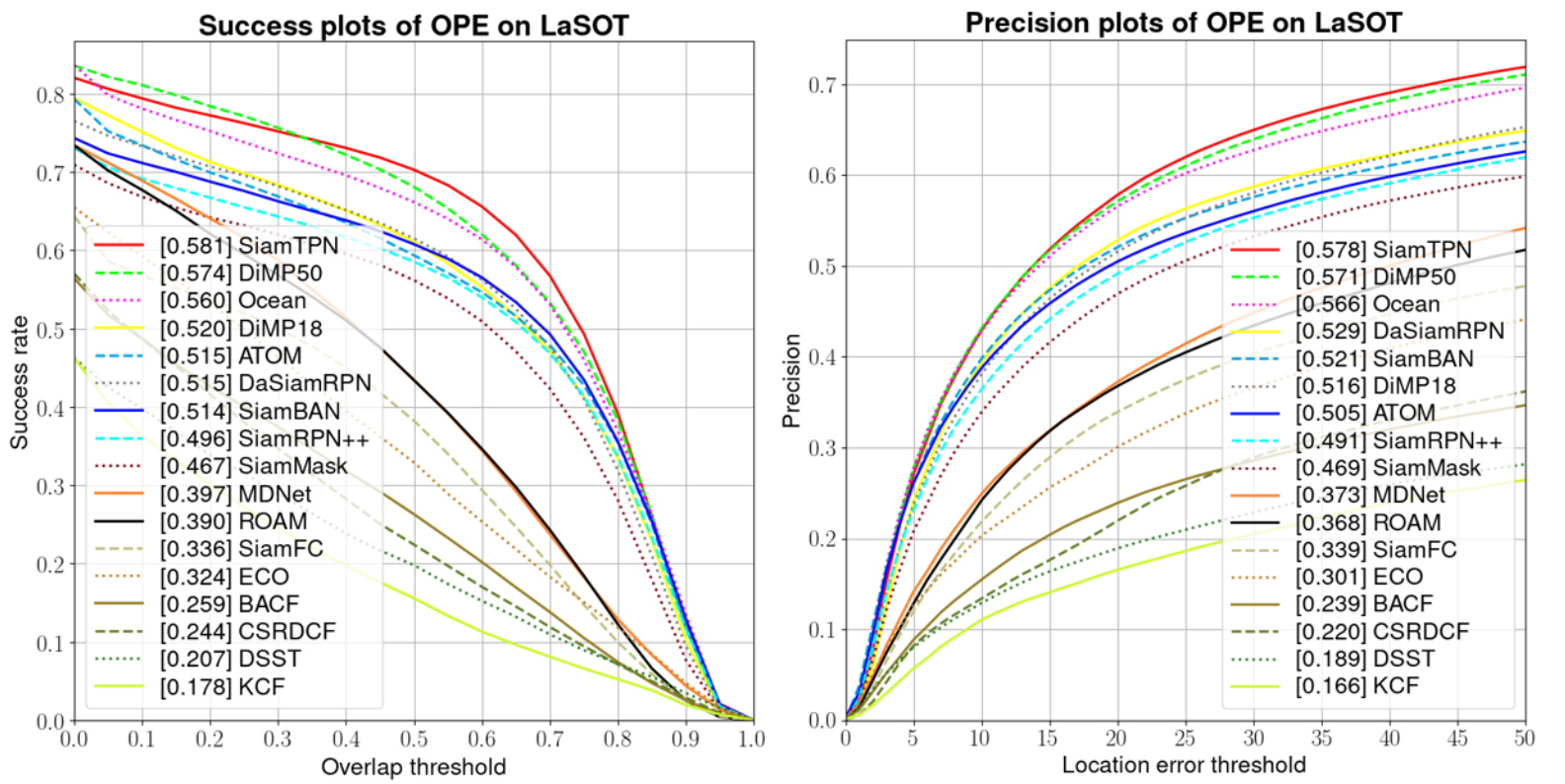}
\end{center}
   \caption{Evaluation results of trackers on LaSOT \cite{lasot}}
\label{lasot}
\end{figure}

In this section, we compare our approach with 22 SOTA trackers. There are 4 anchor-based Siamese methods (SiamRPN~\cite{siamrpn}, SiamRPN++~\cite{siamrpn++}, DaSiamRPN~\cite{dasiamrpn}, HiFT~\cite{hift}), 
5 anchor-free Siamese methods (SiamFC~\cite{siamfc}, SiamBAN~\cite{siamban1}, SiamCar~\cite{siamcar}, SiamFC++~\cite{siamfc++}, Ocean~\cite{ocean}), 
10 DCF based methods (ECO~\cite{ECO}, CCOT \cite{ccot}, KCF~\cite{kcf}, ARCF~\cite{arcf}, BACF~\cite{bacf}, AutoTrack~\cite{autotrack}, CSRDCF \cite{csrdcf}, ROAM \cite{roam}, DiMP \cite{dimp}, ATOM \cite{atom}), 
2 attention based methods (CGACD~\cite{cgacd}, SiamAttn~\cite{siamatt}) and 
1 segmentaion based method, D3S~\cite{d3s}. \\
\textbf{UAV123 \cite{uav123}.} UAV123 is one of the largest UAV tracking benchmarks and adopts success and precision metrics for evaluation. As shown in Table~\ref{uav_result}, all trackers which achieve real-speed time on CPU are based on DCF, which rely on the handcraft features. This becomes a bottleneck of designing high-accuracy trackers. On the other hand, trackers relying on deeper networks like Resnet-50 can achieve high performance but are only applicable on GPU devices. Instead, our SiamTPN runs at real-time speed on the CPU while obtaining SOTA results. Specifically, SiamTPN gains a precision score of 85.8 and an AUC score of 66.04, outperforming the recent SOTA Siamese tracker SiamAttn. For a fair comparison, we develop a variant tracker based on AlexNet. While the AlexNet is not friendly on the CPU end, our tracker could run on the GPU at over 100 FPS while achieving consistent results with SiamRPN++. \\
\textbf{VOT2018 \cite{vot2018} and OTB \cite{otb100}} The VOT2018 dataset consists of 60 sequences with different challenge factors. The performance is compared in terms of EAO (Expected Average Overlap). OTB contains 100 sequences and evaluates performance with AUC score. Table~\ref{vot_otb} shows that our method achieves comparable results with SOTA algorithms on both VOT (second row) and OTB (third row) datasets.\\
\noindent{}\textbf{LaSOT \cite{lasot}.} Figure \ref{lasot} shows our SiamTPN achieves best results on the LaSOT test set, with an AUC score of 58.1 and beats all trackers based on deep Resnet trackers (DiMP, ATOM, OCEAN).\\
\textbf{Got10K~\cite{got10k}} is another large-scale dataset and employs Average Overlap (AO) as measurement. Following the requirement of generic object tracking, there is no overlap in object categories between the training set and test set, which is more challenging and requires a tracker with a powerful generalization ability. We follow their protocol and train the network with a training split. As demonstrated in Figure~\ref{got10k_ao_speed}, SiamTPN achieves a relative $12\%$ higher performance on AO compared with the SOTA Siamese based tracker SiamRPN++ \cite{siamrpn++}. On the other hand, our method exceeds all DCF based trackers while keeping the real-time inference speed on the CPU.

\begin{figure}[htbp]
     \centering
     \begin{subfigure}{0.48\textwidth}
         \centering
         \includegraphics[width=\textwidth]{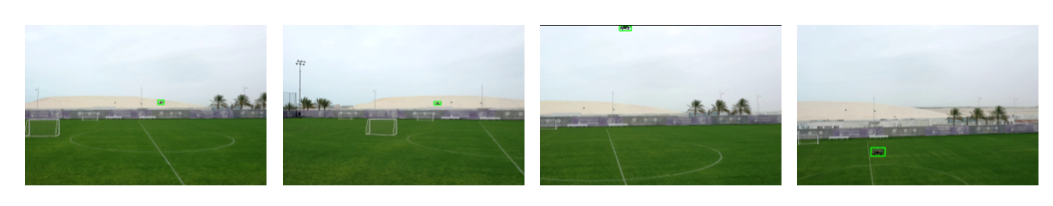}
         \caption{Drone tracking with ground PTZ camera}
         \label{real1}
     \end{subfigure}
     \vfill
     \begin{subfigure}{0.48\textwidth}
         \centering
         \includegraphics[width=\textwidth]{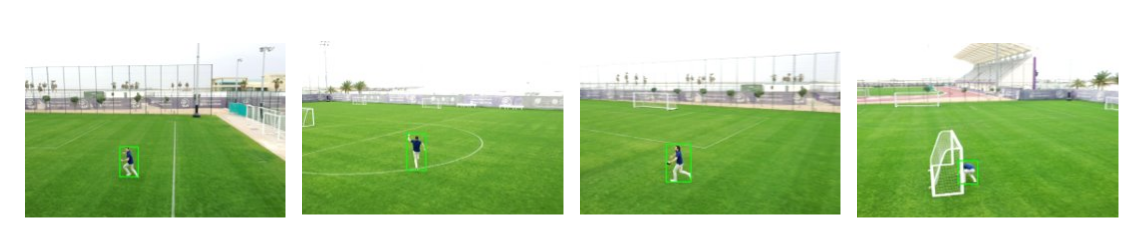}
         \caption{Moving person tracking with a flying drone}
         \label{real2}
     \end{subfigure}
     \vfill
     \begin{subfigure}{0.485\textwidth}
         \centering
         \includegraphics[width=\textwidth]{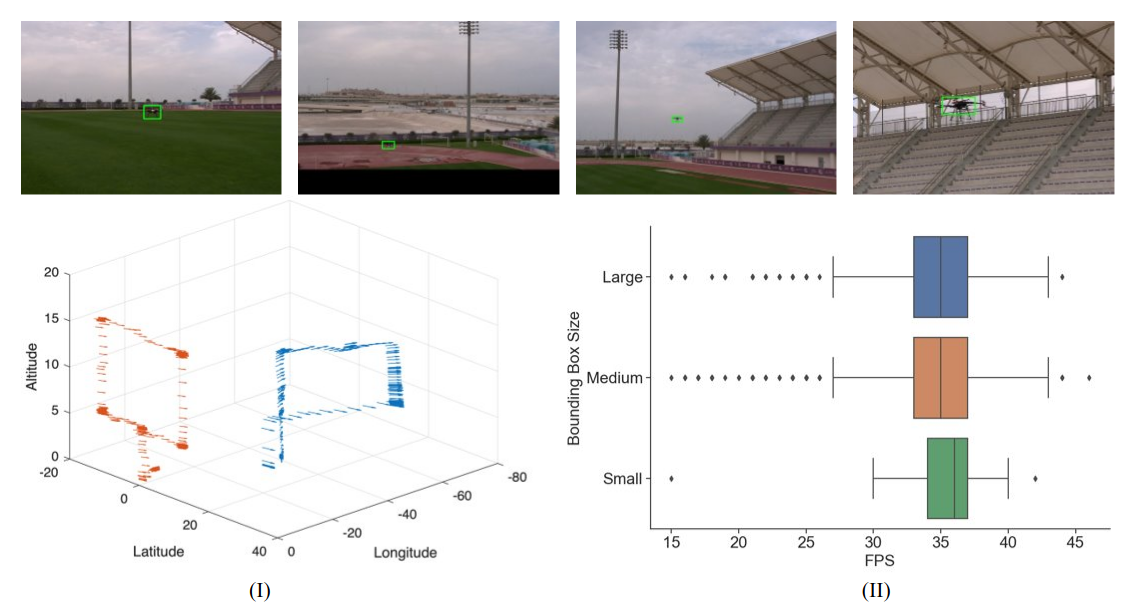}
         \caption{Drone tracking with PTZ camera mounted on a flying drone}
         \label{real3}
     \end{subfigure}
        \caption{Visualization of real-world tracking on drones}
        \label{realworld}
\end{figure}
\subsection{Real World Experimental Test}
In this section, we verify the reliability of the proposed tracker in real-world UAV tracking. The hardware setup consists of a multi-copter UAV, an Embedded PC, a 3 axis Gimbal and a visual PTZ (pan-tilt-zoom) camera. We set up three different tracking scenarios to validate the tracking speed, generalization ability and robustness of SiamTPN. Specifically, the field tests include: (1) drone tracking with a ground stationary PTZ camera, as shown in Figure~\ref{real1}. (2) tracking and following a moving person with a drone and keeping the target within the field of view, as shown in Figure~\ref{real2}. (3) drone (evader) tracking with another drone (pursuer) with PTZ camera embedded, where two drones fly with custom trajectories but the parameters of PTZ camera are adjusted adaptively based on the position of evader, as shown in Figure~\ref{real3}. The position of drones are recorded with two GPS devices and shown in Figure~\ref{real3}(I), where the red~(blue) dots correspond to the pursuer~(evader). Figure~\ref{realworld} shows the precise tracking results obtained under complex environments, exhibiting the robustness and practicability of tracker in real-world applications. We also compare the tracking speed variance under different bounding boxes size. Empirically, we split the bounding boxes into three categories based on pixel numbers, which is small ($<1600$), medium ($<10000$) and large ($>10000$) ones. Figure~\ref{real3}(II) demonstrates the steady inference speed under varies circumstances. 
\section{Conclusion}
In this work, we propose a transformer pyramid network which aggregate semantics from different levels. The local interactions as well as the global dependencies are distilled from the cross attention among pyramid features. A pooling attention is further introduced to prevent the computation overhead. The comprehensive experiments demonstrate that our approach significantly improves the tracking results, while running at real-time speed on the CPU end.
{\small
\bibliographystyle{ieee_fullname}
\bibliography{egbib}

\begin{thebibliography}{10}\itemsep=-1pt

\bibitem{siamfc}
Luca Bertinetto, Jack Valmadre, Joao~F Henriques, Andrea Vedaldi, and Philip~HS
  Torr.
\newblock {Fully-convolutional Siamese networks for object tracking}.
\newblock In {\em European Conference on Computer Vision}, pages 850--865.
  Springer, 2016.

\bibitem{dimp}
Goutam Bhat, Martin Danelljan, Luc~Van Gool, and Radu Timofte.
\newblock Learning discriminative model prediction for tracking.
\newblock In {\em Proceedings of the IEEE International Conference on Computer
  Vision}, pages 6182--6191, 2019.

\bibitem{mosse}
David~S Bolme, J~Ross Beveridge, Bruce~A Draper, and Yui~Man Lui.
\newblock Visual object tracking using adaptive correlation filters.
\newblock In {\em 2010 IEEE computer society conference on computer vision and
  pattern recognition}, pages 2544--2550. IEEE, 2010.

\bibitem{hift}
Ziang Cao, Changhong Fu, Junjie Ye, Bowen Li, and Yiming Li.
\newblock Hift: Hierarchical feature transformer for aerial tracking.
\newblock {\em arXiv preprint arXiv:2108.00202}, 2021.

\bibitem{detr}
Nicolas Carion, Francisco Massa, Gabriel Synnaeve, Nicolas Usunier, Alexander
  Kirillov, and Sergey Zagoruyko.
\newblock End-to-end object detection with transformers.
\newblock In {\em European Conference on Computer Vision}, pages 213--229.
  Springer, 2020.

\bibitem{siamban1}
Zedu Chen, Bineng Zhong, Guorong Li, Shengping Zhang, and Rongrong Ji.
\newblock Siamese box adaptive network for visual tracking.
\newblock In {\em Proceedings of the IEEE/CVF Conference on Computer Vision and
  Pattern Recognition}, pages 6668--6677, 2020.

\bibitem{atom}
Martin Danelljan, Goutam Bhat, Fahad~Shahbaz Khan, and Michael Felsberg.
\newblock Atom: Accurate tracking by overlap maximization.
\newblock In {\em Proceedings of the IEEE Conference on Computer Vision and
  Pattern Recognition}, pages 4660--4669, 2019.

\bibitem{ECO}
Martin Danelljan, Goutam Bhat, Fahad Shahbaz~Khan, and Michael Felsberg.
\newblock Eco: Efficient convolution operators for tracking.
\newblock In {\em Proceedings of the IEEE Conference on Computer Vision and
  Pattern Recognition}, pages 6638--6646, 2017.

\bibitem{dsst}
Martin Danelljan, Gustav H{\"a}ger, Fahad~Shahbaz Khan, and Michael Felsberg.
\newblock Discriminative scale space tracking.
\newblock {\em {IEEE Transactions on Pattern Analysis and Machine
  Intelligence}}, 39(8):1561--1575, 2016.

\bibitem{SRDCF}
Martin Danelljan, Gustav Hager, Fahad Shahbaz~Khan, and Michael Felsberg.
\newblock Adaptive decontamination of the training set: A unified formulation
  for discriminative visual tracking.
\newblock In {\em Proceedings of the IEEE Conference on Computer Vision and
  Pattern Recognition}, pages 1430--1438, 2016.

\bibitem{ccot}
Martin Danelljan, Andreas Robinson, Fahad~Shahbaz Khan, and Michael Felsberg.
\newblock Beyond correlation filters: Learning continuous convolution operators
  for visual tracking.
\newblock In {\em European Conference on Computer Vision}, pages 472--488.
  Springer, 2016.

\bibitem{imagenet}
J. {Deng}, W. {Dong}, R. {Socher}, L. {Li}, {Kai Li}, and {Li Fei-Fei}.
\newblock Imagenet: A large-scale hierarchical image database.
\newblock In {\em 2009 IEEE Conference on Computer Vision and Pattern
  Recognition}, June 2009.

\bibitem{cgacd}
Fei Du, Peng Liu, Wei Zhao, and Xianglong Tang.
\newblock Correlation-guided attention for corner detection based visual
  tracking.
\newblock In {\em Proceedings of the IEEE/CVF Conference on Computer Vision and
  Pattern Recognition}, pages 6836--6845, 2020.

\bibitem{lasot}
Heng Fan, Liting Lin, Fan Yang, Peng Chu, Ge Deng, Sijia Yu, Hexin Bai, Yong
  Xu, Chunyuan Liao, and Haibin Ling.
\newblock Lasot: A high-quality benchmark for large-scale single object
  tracking.
\newblock In {\em Proceedings of the IEEE conference on computer vision and
  pattern recognition}, pages 5374--5383, 2019.

\bibitem{siamcar}
Dongyan Guo, Jun Wang, Ying Cui, Zhenhua Wang, and Shengyong Chen.
\newblock Siamcar: Siamese fully convolutional classification and regression
  for visual tracking.
\newblock In {\em Proceedings of the IEEE/CVF conference on computer vision and
  pattern recognition}, pages 6269--6277, 2020.

\bibitem{resnet}
Kaiming He, Xiangyu Zhang, Shaoqing Ren, and Jian Sun.
\newblock Deep residual learning for image recognition.
\newblock In {\em Proceedings of the IEEE conference on computer vision and
  pattern recognition}, pages 770--778, 2016.

\bibitem{goturn}
David Held, Sebastian Thrun, and Silvio Savarese.
\newblock Learning to track at 100 fps with deep regression networks.
\newblock In {\em European Conference on Computer Vision}, pages 749--765.
  Springer, 2016.

\bibitem{csk}
Joao~F Henriques, Rui Caseiro, Pedro Martins, and Jorge Batista.
\newblock Exploiting the circulant structure of tracking-by-detection with
  kernels.
\newblock In {\em European Conference on Computer Vision}, pages 702--715.
  Springer, 2012.

\bibitem{kcf}
Jo{\~a}o~F Henriques, Rui Caseiro, Pedro Martins, and Jorge Batista.
\newblock High-speed tracking with kernelized correlation filters.
\newblock {\em {IEEE Transactions on Pattern Analysis and Machine
  Intelligence}}, 37(3):583--596, 2014.

\bibitem{got10k}
Lianghua Huang, Xin Zhao, and Kaiqi Huang.
\newblock Got-10k: A large high-diversity benchmark for generic object tracking
  in the wild.
\newblock {\em IEEE Transactions on Pattern Analysis and Machine Intelligence},
  2019.

\bibitem{arcf}
Ziyuan Huang, Changhong Fu, Yiming Li, Fuling Lin, and Peng Lu.
\newblock Learning aberrance repressed correlation filters for real-time uav
  tracking.
\newblock In {\em Proceedings of the IEEE/CVF International Conference on
  Computer Vision (ICCV)}, pages 2891--2900, 2019.

\bibitem{bacf}
Hamed Kiani~Galoogahi, Ashton Fagg, and Simon Lucey.
\newblock Learning background-aware correlation filters for visual tracking.
\newblock In {\em Proceedings of the IEEE Conference on Computer Vision and
  Pattern Recognition}, pages 1135--1143, 2017.

\bibitem{vot2018}
Matej Kristan, Ales Leonardis, Jiri Matas, Michael Felsberg, Roman Pflugfelder,
  Luka ˇCehovin~Zajc, Tomas Vojir, Goutam Bhat, Alan Lukezic, Abdelrahman
  Eldesokey, et~al.
\newblock {The sixth visual object tracking VOT2018 challenge results}.
\newblock In {\em Proceedings of the European Conference on Computer Vision
  (ECCV)}, pages 0--0, 2018.

\bibitem{alexnet}
Alex Krizhevsky, Ilya Sutskever, and Geoffrey~E Hinton.
\newblock Imagenet classification with deep convolutional neural networks.
\newblock {\em Advances in neural information processing systems},
  25:1097--1105, 2012.

\bibitem{pathplanning}
Kuan-Hui Lee and Jenq-Neng Hwang.
\newblock On-road pedestrian tracking across multiple driving recorders.
\newblock {\em IEEE Transactions on Multimedia}, 17(9):1429--1438, 2015.

\bibitem{siamrpn++}
Bo Li, Wei Wu, Qiang Wang, Fangyi Zhang, Junliang Xing, and Junjie Yan.
\newblock {Siamrpn++: Evolution of Siamese visual tracking with very deep
  networks}.
\newblock In {\em Proceedings of the IEEE Conference on Computer Vision and
  Pattern Recognition}, pages 4282--4291, 2019.

\bibitem{siamrpn}
Bo Li, Junjie Yan, Wei Wu, Zheng Zhu, and Xiaolin Hu.
\newblock {High performance visual tracking with Siamese region proposal
  network}.
\newblock In {\em Proceedings of the IEEE Conference on Computer Vision and
  Pattern Recognition}, pages 8971--8980, 2018.

\bibitem{STRCF}
Feng Li, Cheng Tian, Wangmeng Zuo, Lei Zhang, and Ming-Hsuan Yang.
\newblock Learning spatial-temporal regularized correlation filters for visual
  tracking.
\newblock In {\em Proceedings of the IEEE Conference on Computer Vision and
  Pattern Recognition}, pages 4904--4913, 2018.

\bibitem{autotrack}
Yiming Li, Changhong Fu, Fangqiang Ding, Ziyuan Huang, and Geng Lu.
\newblock {AutoTrack: Towards High-Performance Visual Tracking for UAV With
  Automatic Spatio-Temporal Regularization}.
\newblock In {\em Proceedings of the IEEE/CVF Conference on Computer Vision and
  Pattern Recognition (CVPR)}, pages 11920--11929, 2020.

\bibitem{fpn}
Tsung-Yi Lin, Piotr Doll{\'a}r, Ross Girshick, Kaiming He, Bharath Hariharan,
  and Serge Belongie.
\newblock Feature pyramid networks for object detection.
\newblock In {\em Proceedings of the IEEE conference on computer vision and
  pattern recognition}, pages 2117--2125, 2017.

\bibitem{coco}
Tsung-Yi Lin, Michael Maire, Serge Belongie, James Hays, Pietro Perona, Deva
  Ramanan, Piotr Doll{\'a}r, and C~Lawrence Zitnick.
\newblock Microsoft coco: Common objects in context.
\newblock In {\em European Conference on Computer Vision}, pages 740--755.
  Springer, 2014.

\bibitem{panet}
Shu Liu, Lu Qi, Haifang Qin, Jianping Shi, and Jiaya Jia.
\newblock Path aggregation network for instance segmentation.
\newblock In {\em Proceedings of the IEEE conference on computer vision and
  pattern recognition}, pages 8759--8768, 2018.

\bibitem{adamw}
Ilya Loshchilov and Frank Hutter.
\newblock Decoupled weight decay regularization.
\newblock {\em arXiv preprint arXiv:1711.05101}, 2017.

\bibitem{d3s}
Alan Lukezic, Jiri Matas, and Matej Kristan.
\newblock D3s - a discriminative single shot segmentation tracker.
\newblock In {\em CVPR}, 2020.

\bibitem{csrdcf}
Alan Lukezic, Tomas Vojir, Luka ˇCehovin~Zajc, Jiri Matas, and Matej Kristan.
\newblock Discriminative correlation filter with channel and spatial
  reliability.
\newblock In {\em Proceedings of the IEEE Conference on Computer Vision and
  Pattern Recognition}, pages 6309--6318, 2017.

\bibitem{shufflenetv2}
Ningning Ma, Xiangyu Zhang, Hai-Tao Zheng, and Jian Sun.
\newblock {Shufflenet v2: Practical guidelines for efficient CNN architecture
  design}.
\newblock In {\em Proceedings of the European Conference on Computer Vision
  (ECCV)}, pages 116--131, 2018.

\bibitem{uav123}
Matthias Mueller, Neil Smith, and Bernard Ghanem.
\newblock {A benchmark and simulator for UAV tracking}.
\newblock In {\em European Conference on Computer Vision}, pages 445--461.
  Springer, 2016.

\bibitem{trackingnet}
Matthias Muller, Adel Bibi, Silvio Giancola, Salman Alsubaihi, and Bernard
  Ghanem.
\newblock Trackingnet: A large-scale dataset and benchmark for object tracking
  in the wild.
\newblock In {\em Proceedings of the European Conference on Computer Vision
  (ECCV)}, pages 300--317, 2018.

\bibitem{dat}
Horst Possegger, Thomas Mauthner, and Horst Bischof.
\newblock In defense of color-based model-free tracking.
\newblock In {\em Proc. IEEE Conference on Computer Vision and Pattern
  Recognition (CVPR)}, 2015.

\bibitem{giou}
Hamid Rezatofighi, Nathan Tsoi, JunYoung Gwak, Amir Sadeghian, Ian Reid, and
  Silvio Savarese.
\newblock Generalized intersection over union: A metric and a loss for bounding
  box regression.
\newblock In {\em Proceedings of the IEEE/CVF Conference on CVPR}, pages
  658--666, 2019.

\bibitem{mobilenetv2}
Mark Sandler, Andrew Howard, Menglong Zhu, Andrey Zhmoginov, and Liang-Chieh
  Chen.
\newblock Mobilenetv2: Inverted residuals and linear bottlenecks.
\newblock In {\em {Proceedings of the IEEE Conference on CVPR}}, pages
  4510--4520, 2018.

\bibitem{vgg}
Karen Simonyan and Andrew Zisserman.
\newblock Very deep convolutional networks for large-scale image recognition.
\newblock {\em arXiv preprint arXiv:1409.1556}, 2014.

\bibitem{surveillance}
Siyu Tang, Mykhaylo Andriluka, Bjoern Andres, and Bernt Schiele.
\newblock Multiple people tracking by lifted multicut and person
  re-identification.
\newblock In {\em Proceedings of the IEEE Conference on CVPR}, pages
  3539--3548, 2017.

\bibitem{Tsoukalas}
Athanasios Tsoukalas, Daitao Xing, Nikolaos Evangeliou, Nikolaos Giakoumidis,
  and Anthony Tzes.
\newblock Deep learning assisted visual tracking of evader-{UAV}.
\newblock In {\em 2021 International Conference on Unmanned Aircraft Systems
  (ICUAS)}, pages 252--257. IEEE, 2021.

\bibitem{transformer}
Ashish Vaswani, Noam Shazeer, Niki Parmar, Jakob Uszkoreit, Llion Jones,
  Aidan~N Gomez, {\L}ukasz Kaiser, and Illia Polosukhin.
\newblock Attention is all you need.
\newblock In {\em Advances in neural information processing systems}, pages
  5998--6008, 2017.

\bibitem{otb100}
Yi Wu, Jongwoo Lim, and Ming-Hsuan Yang.
\newblock Online object tracking: A benchmark.
\newblock In {\em Proceedings of the IEEE Conference on Computer Vision and
  Pattern Recognition}, pages 2411--2418, 2013.

\bibitem{siamfc++}
Yinda Xu, Zeyu Wang, Zuoxin Li, Ye Yuan, and Gang Yu.
\newblock Siamfc++: Towards robust and accurate visual tracking with target
  estimation guidelines.
\newblock In {\em AAAI}, pages 12549--12556, 2020.

\bibitem{roam}
Tianyu Yang, Pengfei Xu, Runbo Hu, Hua Chai, and Antoni~B Chan.
\newblock {ROAM: Recurrently Optimizing Tracking Model}.
\newblock In {\em CVPR}, 2020.

\bibitem{siamatt}
Yuechen Yu, Yilei Xiong, Weilin Huang, and Matthew~R Scott.
\newblock Deformable siamese attention networks for visual object tracking.
\newblock In {\em Proceedings of the IEEE/CVF Conference on Computer Vision and
  Pattern Recognition}, pages 6728--6737, 2020.

\bibitem{ocean}
Zhipeng Zhang, Houwen Peng, Jianlong Fu, Bing Li, and Weiming Hu.
\newblock Ocean: Object-aware anchor-free tracking.
\newblock In {\em The European Conference on Computer Vision (ECCV)}, August
  2020.

\bibitem{dasiamrpn}
Zheng Zhu, Qiang Wang, Bo Li, Wei Wu, Junjie Yan, and Weiming Hu.
\newblock {Distractor-aware Siamese networks for visual object tracking}.
\newblock In {\em Proceedings of the European Conference on Computer Vision
  (ECCV)}, pages 101--117, 2018.

\end{thebibliography}
}

\end{document}